\documentclass{article}

\usepackage{arxiv}

\usepackage[utf8]{inputenc} % allow utf-8 input
\usepackage[T1]{fontenc}    % use 8-bit T1 fonts
\usepackage{hyperref}       % hyperlinks
\usepackage{url}            % simple URL typesetting
\usepackage{booktabs}       % professional-quality tables
\usepackage{amsfonts}       % blackboard math symbols
\usepackage{nicefrac}       % compact symbols for 1/2, etc.
\usepackage{microtype}      % microtypography

\usepackage{graphicx}
\usepackage{doi}
\usepackage{verbatim}
\usepackage{xcolor,colortbl}
\usepackage{tabularx}
\usepackage{miller}
\usepackage{layouts}
\usepackage{svg}
\usepackage{mathtools}
\usepackage{subfig}
\usepackage{wrapfig}
\usepackage{epstopdf}

\definecolor{LightYellow}{rgb}{1,1,0.878}
\newcolumntype{a}{>{\columncolor{LightYellow}}l}

\title{MaterioMiner --- An ontology-based text mining dataset for extraction of process-structure-property entities}

\author{ \href{https://orcid.org/0000-0002-0916-5990}{\includegraphics[scale=0.06]{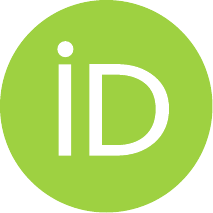}\hspace{1mm}Ali Riza Durmaz}$\dag$ \\
	Group of Meso and Micromechanics\\
	Fraunhofer Institute for Mechanics of Materials IWM\\
	Freiburg im Breisgau, 79108, Germany \\
	\texttt{ali.riza.durmaz@iwm.fraunhofer.de} \\
	\And
	\href{https://orcid.org/0000-0003-0744-8855}{\includegraphics[scale=0.06]{orcid.pdf}\hspace{1mm}Akhil Thomas}$\dag$ \\
	Group of Meso and Micromechanics\\
	Fraunhofer Institute for Mechanics of Materials IWM\\
	Freiburg im Breisgau, 79108, Germany \\
	\texttt{akhil.thomas@iwm.fraunhofer.de} \\
	\And
	\href{https://orcid.org/0000-0002-1256-7261}{\includegraphics[scale=0.06]{orcid.pdf}\hspace{1mm}Lokesh Mishra} \\
	IBM Research\\
	Rüschlikon, 8803, Switzerland \\
        \texttt{mis@zurich.ibm.com} \\
    \And
	Rachana Niranjan Murthy \\
	Group of Meso and Micromechanics\\
	Fraunhofer Institute for Mechanics of Materials IWM\\
	Freiburg im Breisgau, 79108, Germany \\
 \And
	Thomas Straub \\
	Group of Meso and Micromechanics\\
	Fraunhofer Institute for Mechanics of Materials IWM\\
	Freiburg im Breisgau, 79108, Germany \\
}

\renewcommand{\shorttitle}{MaterioMiner --- A text mining dataset for extraction of materials entities}

\hypersetup{
pdftitle={A template for the arxiv style},
pdfsubject={q-bio.NC, q-bio.QM},
pdfauthor={David S.~Hippocampus, Elias D.~Striatum},
pdfkeywords={First keyword, Second keyword, More},
}

\begin{document}
\maketitle

\begin{abstract}

While large language models learn sound statistical representations of the language and information therein, ontologies are symbolic knowledge representations that can complement the former ideally. Research at this critical intersection relies on datasets that intertwine ontologies and text corpora to enable training and comprehensive benchmarking of neurosymbolic models. We present the \textit{MaterioMiner dataset} and the linked \textit{materials mechanics ontology} where ontological concepts from the mechanics of materials domain are associated with textual entities within the literature corpus. Another distinctive feature of the dataset is its eminently fine-granular annotation. Specifically, 179 distinct classes are manually annotated by three raters within four publications, amounting to a total of 2191 entities that were annotated and curated. Conceptual work is presented for the symbolic representation of causal composition-process-microstructure-property relationships. We explore the annotation consistency between the three raters and perform fine-tuning of pre-trained models to showcase the feasibility of named-entity recognition model training.  Reusing the dataset can foster training and benchmarking of materials language models, automated ontology construction, and knowledge graph generation from textual data.

\end{abstract}

\footnote{$\dag$: These authors contributed equally to this work}

% keywords can be removed

\section{Background \& Summary}
We present a rich annotated dataset and a new ontology targeting the materials mechanics domain with a focus on materials fatigue. The outlined dataset and ontology facilitate the extraction of detailed information on materials compositions, underwent processing and experimentation, resulting defect distributions, and properties from unstructured textual data. This entails both, the training and benchmarking of corresponding text mining models. While for general domains, data mining methodologies are in place to extract fine-grained information from publications and compose knowledge graphs with decent data quality \cite{buscaldi2019mining, khorashadizadeh2023exploring}, such techniques are only nascent in the material science and engineering (MSE) field \cite{yang2022pcmsp, buehler2023mechgpt}. In particular, the recognition and extraction of detailed, domain-specific information are still challenging. In many proposed pipelines, the information extraction happens in an unconstrained fashion. However, in many cases, it would be desirable to extract domain knowledge such that it conforms to an existing knowledge base or ontological framework, e.g. some mid-level materials ontology. In doing so, deduced information can be aligned with existing knowledge graph repositories which were constructed meticulously using analogous semantic annotation. Furthermore, complying with a provided data schema allows straightforward querying. Along those lines, current efforts try to fine-tune large language models (LLMs) using an ontological data schema to obtain domain-conformal linked data at inference time \cite{LawrenceText2Graph}.  

The presented semantically annotated dataset can also be used to train specialized language models. Specifically, a token classification task called named-entity recognition (NER) for a wide range of material entities can be addressed directly. Given a set of named entities of interest, NER aims to detect and categorize important information in text into one of these named entities. In our work, we distinguish between coarse-granular NER (CG-NER) and fine-granular NER (FG-NER), which address the recognition of high- and low-level concepts in an accompanying application ontology, respectively. While we link text entities to a knowledge base, we do not consider this a named-entity linking (NEL) task, since our knowledge base does not contain individuals/instances of specific classes that are disambiguated. Furthermore, the proposed application ontology lays the groundwork for unified relation extraction (RE). Claiming that multiple-step pipelines that perform NER followed by RE are unfavorable in terms of error propagation, the REBEL model\cite{cabot2021rebel} proposes performing direct triple extraction using a seq2seq BART model. The provided data can lay the groundwork for fine-tuning such triple extraction techniques with relevant relations in the materials mechanics domain.  

Prospectively, the data is expected to foster the extraction of causal composition-process-microstructure-property (CPMP) relationships \cite{hattrick2016perspective} and population of materials databases from textual data. Such text mining permits exploring mechanisms and driving forces for specific degradation phenomena. Materials mechanics has many controversies, e.g., the underlying mechanism of intergranular crack growth \cite{mazanova2022mechanism}. To explore such controversies and resolve inconsistencies, extracting detailed information about the process history from textual data and applying deductive, inductive, abductive, and counterfactual reasoning is promising. Ontologies, due to their foundation in description and first-order logic, support computational reasoning capabilities. Aside from that, exploiting information in textual data can complement well-structured numerical and categorical features and foster better fidelity in prediction tasks. For instance, the added value of embedding textual information along with numerical and categorical variables for predicting the pitting potential (regression task) was demonstrated \cite{sasidhar2023enhancing}.

Large language models such as GPT or Llama variants recently gained a lot of attention and excel at understanding nuances of natural language. However, repurposing such foundation models for specialized tasks, even in domain-unspecific settings, to date often falls short of specialized models \cite{qin2023chatgpt}, which also holds for NER \cite{wang2023gpt}. Furthermore, logical reasoning is not necessarily a strength of LLMs. In this sense, language models are complementary to ontologies, which formalize a domain using axioms constructed from explicitly defined classes, individuals, and object/data properties (relations). Thus ontologies can be seen as abstract, structured, standardized, and formal knowledge models of domains. These characteristics render ontologies and models based on them ideal for interpretability and retracing decision-making. Ontologies rely on description logic to model class hierarchy as well as other axioms/constraints. Thus, the integration of language models and explicit, symbolic knowledge in ontologies, which falls into the category of neurosymbolic artificial intelligence \cite{de2020statistical}, can permit logical inference of new knowledge based on the existing domain models. For instance, LLM agents that can additionally perform triple validation using shape constraint language (SHACL) and ontological reasoning engines can be envisioned. This can potentially culminate in increased accuracy and generalization of combined models. Furthermore, GPT variants were reported to gradually drift and degrade on some reasoning tasks \cite{chen2023chatgpt} despite chain-of-thought prompting \cite{wei2022chain}. This can potentially be alleviated through infusing ontological constraints as this might not only improve the model's reasoning capabilities but also stability. Especially given the steadily increasing pace of research and overpublishing \cite{akbashev2023tackling}, automated bottom-up methodologies that extract validated knowledge and data from literature sources and perform plausibility checks are needed to ensure the effective assimilation of literature information. Furthermore, foundation language models trained on arbitrary data perform well at general tasks but struggle when it comes to domain-specific problems \cite{xie2023darwin}. Ontologies can alleviate this issue by introducing relevant relationships and constraints, e.g. about the materials domain. Thus, through the combination of both approaches unstructured and structured data can be harnessed jointly.

\begin{table}[h!t]
\centering
\begin{tabularx}{\textwidth}{|X|X|X|X|X|X|}
\hline
\textbf{Name} & \textbf{(SOFC)-Exp corpus}\cite{friedrich2020sofc} & \textbf{MaterialsBERT}\newline{\cite{materialsbert}} & \textbf{Matscholar NER dataset}\cite{weston2019named} & \textbf{PcMSP dataset}\cite{yang2022pcmsp} & \textbf{MaterioMiner dataset (ours)}\\
\hline
\textbf{Domain} & Solid oxide fuel cells & Organic materials & Materials synthesis & Polycrystalline Materials Synthesis & Material mechanics\\
\hline
\textbf{Tasks} & NER, topic classification, and slot filling & NER & NER & NER, RE, sentence classification & NER (different granularities)\\
\hline
\textbf{Annotated data quantity} & 45 publications & 750 abstracts & 800 abstracts  & experimental paragraphs of 305 publications& 4 publications \\
\hline
\textbf{Tokens [count]} & 288,373 & 24,578 & 111,380 & 66,944 & 12,155 \\
\hline
\textbf{Annotated tokens [\%]} & 2\% & 14\% & 20\% & 22\% & \textbf{27\%} \\
\hline
\textbf{\# Annotated classes} & 4 & 8 & 7 & 13 & \textbf{27/179*}\\
\hline
\textbf{Entity types} & material, value, device, and experiment & polymer, polymer class, property name, property value, monomer,
organic material, inorganic material, and material amount & inorganic material, symmetry/phase label, sample descriptor, material property, material application, synthesis method, and characterization method & brand, descriptor, device, material(-intermedium, -others, -recipe, -target), operation, property(-pressure, -rate, -temperature, -time), value & For a list of annotated classes, see Figure \ref{fig:Class_distribution}b and published FG-NER CoNLL file for coarse and fine-granular NER, respectively.\\
\hline
\end{tabularx}
\caption{A comparison of currently available datasets in the broader materials science domain. The datasets are vaguely ranked from left to right by the annotation density and number of classes which jointly can be seen as a measure of the degree of detail. Foppiano et al. use a notable NER dataset for superconductor materials\cite{foppiano2021supermat}. However, in their case, the NER data could not be published due to copyright constraints. The asterisk indicates the number of annotated ontological classes in the fine-granular NER case.}
\label{tab:dataset_comparison}
\end{table}

To enable research at the intersection between ontological knowledge representations, and language models within the materials field, we present a dataset that links a custom materials mechanics application ontology with corresponding entities observed in texts. The semantically annotated data is expected to promote ontology development and linked data extraction from literature sources. Furthermore, we anticipate the development of advanced language models that exploit the ontology in conjunction with reasoning engines for better text generation, disambiguation, and other tasks. The accompanying materials ontology builds on the PMDco ontology \cite{bayerlein2023pmd, PMDco2}, a midlevel materials ontology which in turn borrows from the PROV ontology \cite{lebo2013prov}. 

There are some related textual materials science datasets published which are summarized in Table \ref{tab:dataset_comparison}. While most available datasets cover specific subdomains, e.g. fuel cells, superconductors, organic/polycrystal material synthesis, and are focused on high-level information, our dataset, whilst capturing very detailed information, refers to an ontology that is largely MSE domain-agnostic. Thus, the proposed ontology provides a framework to map equivalent and related concepts from different datasets to each other to normalize and consolidate the data. At the same time, ontologies offer the possibility for data standardization where each concept (and consequentially each named entity) is explicitly defined.

\section{Methods}
\label{sec:headings}

\begin{figure}[h!t]
\centering
\includegraphics[width=\linewidth]{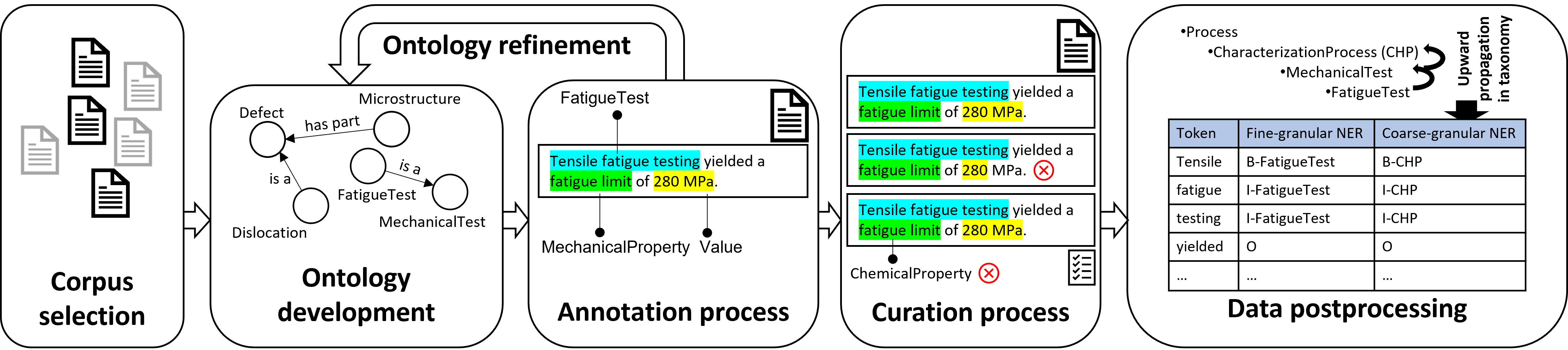}
\caption{Schematic showing the methodology used to generate the presented dataset. The ontology was refined manually in an iterative fashion to permit thorough annotation of materials science-related scholarly articles.}
\label{fig:workflow}
\end{figure}

An overview of the procedure applied for producing the data is illustrated in Figure \ref{fig:workflow}. An application ontology based on existing materials ontologies was built which permits the materials domain-agnostic annotation of textual entities. A literature corpus that covers the materials mechanics domain was collected and annotated using the application ontology, which in return was gradually refined in the process. After annotation, a curation process was performed to harmonize the data culminating in a validated dataset for training and benchmarking of neurosymbolic models. The dataset intertwines an ontological knowledge base with textual data literals from the corpus. Data post-processing involving the ontology and the annotated articles was conducted to demonstrate the advantage of the ontology-integrated approach culminating in two NER datasets with distinct granularity.  Resulting fine-granular and coarse-granular NER datasets can act as benchmark NER datasets within the fatigue domain.

In the following, a notation will be employed where ontological entities are indicated by prefixes. Classes and object properties exhibit regular and italic font styles, respectively.

\subsection*{Ontology development}
\label{sec:onto-development}
% Dependencies of the ontology
The published dataset links to a tailored materials ontology, called \textit{material mechanics ontology} (prefix: \textit{mm}) and published as version v1.0.0, which is mapped on the PMDco 2.0.4 ontology \cite{PMDco2}. PMDco (prefix: \textit{pmd}) is a midlevel materials ontology that models fundamental concepts in materials science and engineering (MSE) and in turn borrows from the PROV (prefix: \textit{prov}) top-level ontology \cite{lebo2013prov}. The PMDco ontology aims to provide a framework for MSE to render derived domain and application ontologies compatible and applications building on those ontologies interoperable \cite{bayerlein2023pmd}. However, mapping the material mechanics ontology onto other upper materials ontologies such as the elementary materials modeling ontology (EMMO)\cite{horsch2020semantic} is equally feasible. Furthermore, our materials mechanics ontology utilizes concepts representing material properties from the MP-SCHM ontology which is a part of the semantic materials, manufacturing, and design project \cite{PaulSEMMD}. 

% Overview and expectation management
The materials mechanics ontology covers some fairly general MSE concepts such as descriptions of crystallographic defects and microstructural entities that could prospectively be merged into PMDco or other upper materials ontologies. Furthermore, concepts related to the materials fatigue subdomain are also heavily incorporated since fatigue represented the use case for which the textual data annotation was performed. While the materials mechanics ontology, in its current version, does not claim to model object property constraints on classes and other axioms comprehensively, we attempt to conceptualize how CPMP relationships in materials mechanics (e.g. causation or influence) can be modeled through object properties existing in aforementioned upper ontologies or object properties introduced in our ontology. The application ontology before PMDco import and mapping consists of 427 classes distributed over four hierarchy levels. In the published state, the proposed ontology does not cover individuals/instances such as existing specimens, i.e. contains only T-Box statements as opposed to A-Box ones. However, the ontological knowledge base, given the international resource identifiers (IRIs) therein, prospectively provides a framework that readily supports NEL, i.e. disambiguating entities at an instance level, as well.

% How is the ontology built 
Rather than applying one of the known structured approaches to ontology design building on competency questions\cite{poveda2022lot, peroni2016samod}, an iterative schema was applied in which the scope of the ontology was gradually refined to incorporate relevant concepts to permit extensive annotation of scholarly articles in MSE. In practice, the initial ontology was automatically generated starting from a glossary using the `excelparser' module from the EMMOntoPy python package~\cite{jesper_friis_2024_10573000}, and the iterative refinement was performed in Protégé \cite{musen2015protege} by manually adding and shifting classes. The ontology refinement considered an extended literature corpus comprising 51 publications to attain an ontology that can represent many entities relevant to MSE. As a consequence of this integrated approach, the annotators were heavily involved during the ontology development stage which not only improved the quality of the ontology but also led to a better understanding of the ontology's scope, perspectives, and class definitions. This is necessary when the objective is an ontology-conformal and fine-granular annotation that preserves most information.

\begin{figure}[ht]
\centering
%\includesvg[width=\linewidth]{images/cpsp_material_mechanics_ontology_adjusted.svg}
\includegraphics[width=\linewidth]{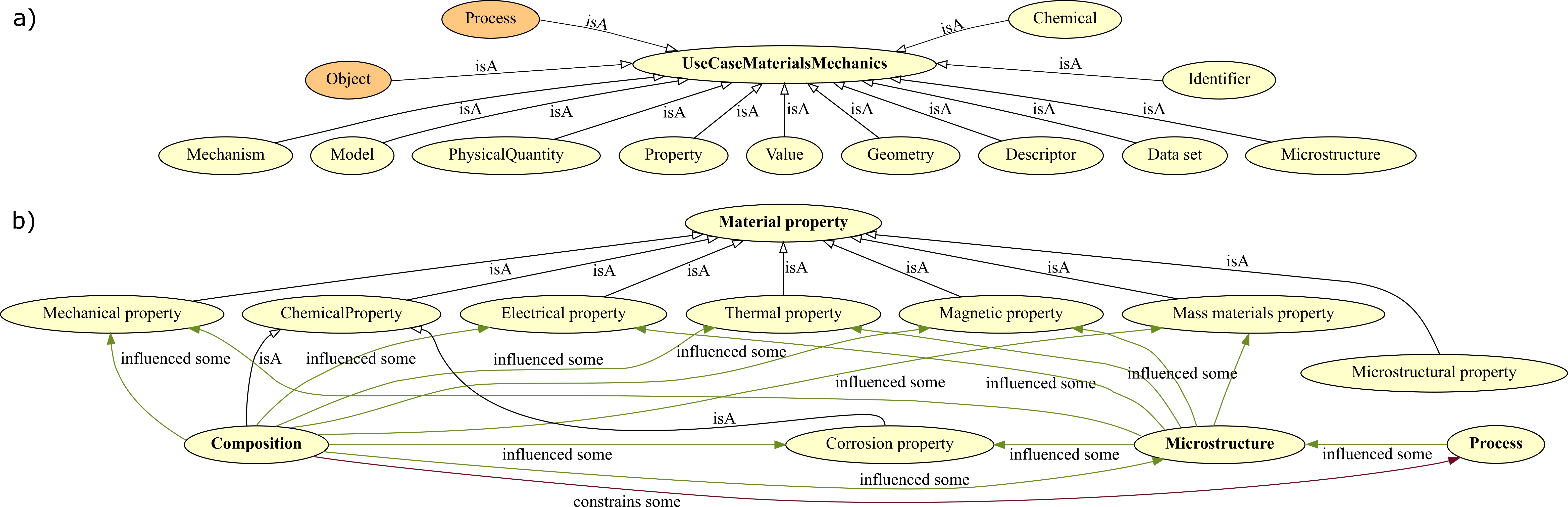}
\caption{a) The top-level structure of the proposed materials mechanics ontology covers the most common entities in materials science. b) General concepts capturing composition-processing-microstructure-property relationships. Subclass relations are displayed in black while other object properties are assigned colors. Ontology prefixes are discarded for clear visual display. Instead, classes to which an equivalent class exists in PMDco are visualized with an orange background. In such cases, an \textit{owl:equivalent} relation was used to express equivalence.}
\label{fig:toplevel}
\end{figure}

The top-level structure of the proposed application ontology is depicted in Figure \ref{fig:toplevel}a and spans a variety of relevant classes and concepts. For instance, the mm:PhysicalQuantity class supports the annotation of labels, i.e. names, and symbols of physical quantities while mm:Value is used for literals (and their unit). The material mechanics ontology aims to enable representations of mechanistic causation across material size scales. The general CPMP structure is illustrated in Figure \ref{fig:toplevel}b and more detailed views illustrating the modeling of crystallographic defects and damage are provided in Figure \ref{fig:defects} and \ref{fig:damage}, respectively. Materials properties are subdivided into different property groups and explicitly modeled as classes rather than relations (data or object properties). This allows adding domain knowledge as property restrictions connecting related concepts, for example, that mm:GrainBoundaryHardening is a sufficient cause of mm:TensileYieldStrength. The composition is considered an inherent, multi-valued chemical property of a material.

Adjacent efforts in terms of microstructure modeling attempted to represent the crystallographic information framework (CIF) using ontologies, resulting in the so-called CIF core ontology \cite{hall2016implementation}. The corresponding concepts were then also adopted by EMMO establishing the crystallography domain ontology (CDO) v0.1.0 \cite{rickard_armiento_2023_7966654}. Moreover, the microstructure domain ontology (MDO) v0.1.0 \cite{jesper_friis_2023_7966660} of EMMO which imports the CDO and is currently being developed exhibits a related scope to our effort. Similarly, Ihsan et al. \cite{ihsan2021steps, ihsan2023diso} developed an ontology called `DISO' focussed on dislocations.

Specifically, our material mechanics application ontology has a significant overlap with the aforementioned EMMO modules when classes used for microstructure modeling are concerned. A common means to distinguish different crystallographic defects is their dimensionality, i.e. point, line, planar, and volume defects. We opted for a similar taxonomy, as also used in EMMO's microstructure domain ontology (MDO), as indicated in Figure \ref{fig:defects}. However, while these modules of EMMO adhere closer to the data level by adopting concepts such as `PixelGrid', we largely disregard these concepts that relate to binary data and put more emphasis on comprehensiveness for concepts that are common in MSE literature.  

% Association, correlation and causation \cite{altman2015points}
% \cite{sawesi2022representation}

\subsubsection*{Modeling of causal composition-processing-microstructure-property relationships}

In terms of object properties, the ontology contains subclass relations (taxonomy) and other arbitrary relations as constraints between classes. The \textit{isA} relation shown in the figures represents the subclass relation \textit{rdfs:SubClassOf} which causes the inheritance of class constraints throughout the hierarchy levels. Subsequently, we propose how different object properties can be used to model relationships such as association, correlation, and causation. We attempt to underpin this modeling proposition with examples from the materials mechanics domain. The proposed object properties should permit a high degree of expressivity. For instance, expressive semantic representations are essential for reasoning, causal deep learning, and knowledge graph embedding methods. To the best of our knowledge, this is the first attempt at symbolically representing such relationships within the CPMP chain through ontologies. In the biomedical domain, Sawesi et al. surveyed a variety of modeling approaches for causal relationships \cite{sawesi2022representation}.

A general relationship that implies that the subject holds information about the object is proposed in \textit{mm:associatedWith}. Here, we align with the viewpoint of Altman and Krzywinski \cite{altman2015points}. A sub-property would be \textit{mm:correlatedWith} which additionally implies an increasing/decreasing trend within the dependent and independent variable (or between the linked subject and object). While correlation does not imply causation, the latter, represented by \textit{mm:causeOf}, is still considered a sub-property of \textit{mm:associatedWith}. \textit{mm:causeOf} has an intersection with \textit{mm:correlatedWith} and is applied whenever a subject is partly responsible for the object. Further, we model two sub-properties of \textit{mm:causeOf}, namely \textit{mm:necessaryCauseOf} and \textit{mm:sufficientCauseOf} \cite{befani2012models}. Former addresses cases where the presence of the object implies the prior occurrence of the subject. An example could be fatigue cracks which imply that cyclic loads occured, however, cracks do not necessarily initiate under the same cyclic loads given the multitude of other influence factors. On the other hand, \textit{mm:sufficientCauseOf} indicates cases where the presence of the subject implies the subsequent occurrence of the object (i.e., no other conditions need to be fulfilled). The opposite does not apply though, since the object can also be attained by other means. For example, loads above a certain threshold in many coarse-grained alloys are sufficient to cause deformation twinning \cite{zhu2013grain}. However, twinning can be also caused during thermo-mechanical processing (annealing twins).

Another relation that is commonly used as object property constraint is \textit{prov:influenced}, which is a directed relationship that indicates a generic capacity of the subject to affect the character, development, or behavior of the object. We consider it a super-property of \textit{mm:causeOf}, since \textit{prov:influenced}, as opposed to \textit{mm:causeOf}, additionally entails the more specific sub-property \textit{mm:constrains} and its sub-properties \textit{mm:hardConstrains} and \textit{mm:softConstrains}. These are used to signify that the subject imposes some bounds on the object. For instance, the recycling-induced elevated presence of copper in steel asserts some soft bounds (\textit{mm:softConstrains}) on its processing window to avoid embrittlement and hot shortness \cite{shibata2002suppression, melford1980influence}. In contrast, a load cell built into a mechanical testing setup asserts hard bounds (\textit{mm:hardConstrains}) to the measurable force. Another sub-property of \textit{prov:influenced} is \textit{prov:generated} which is used to relate prov:Activities to produced prov:Entities. In the materials mechanics ontology, this was used to describe that the Frank-Read source mechanism produces dislocations (mm:FrankReadSource \textit{prov:generated} some mm:Dislocation). In the object property restrictions added to classes, the existential restriction `some' was widely used for high-level concepts. For instance, the mm:Microstructure \textit{prov:influenced some} mm:MechanicalProperty restriction indicates that all individuals of mm:Microstructure (and by extension, all subclasses of it) are connected to at least one individual of mm:MechanicalProperty (e.g. yield strength) through the \textit{prov:influenced} object property. Note that the absence of such triples would not cause reasoning errors due to the OWL's open-world assumption \cite{keet2013open}.

\begin{figure}[ht]
\centering
%\includesvg[width=\linewidth]{images/defects_material_mechanics_ontology_adjusted.svg}
\includegraphics[width=\linewidth]{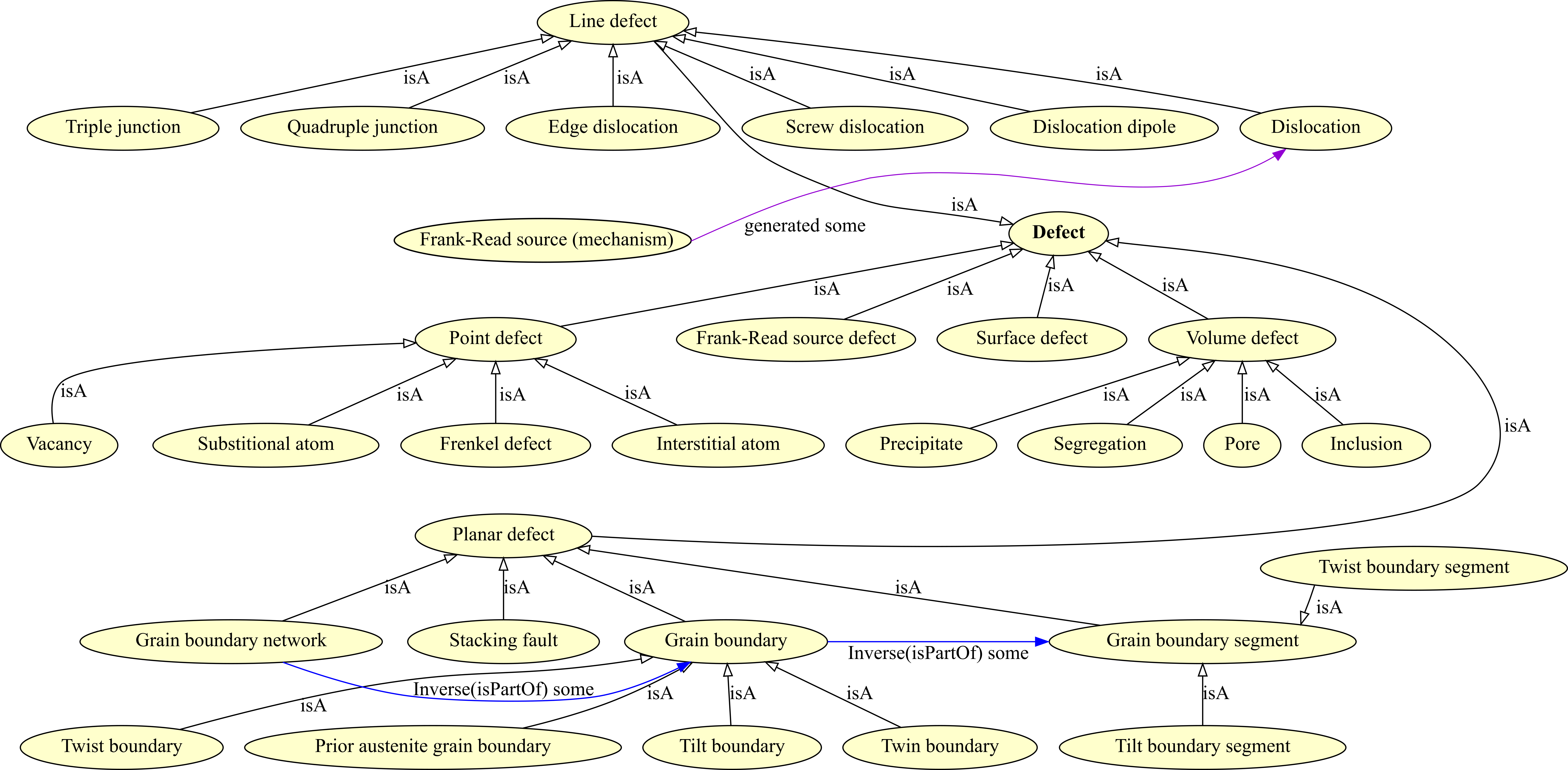}
\caption{The hierarchy for defects is shown using point, line, planar, surface, and volume defects as superclasses for common crystallographic defects. Specific configurations such as Frank-Read source defects, i.e. pinned dislocations are also considered defects that can cause dislocation multiplication through the Frank-Read source mechanism. A text `dislocation' is annotated as the equivalently named ontology class and additionally propagated upwards to the mm:Defect class using the mm:Dislocation $\xrightarrow[]{\text{isA}}$ mm:LineDefect $\xrightarrow[]{\text{isA}}$ mm:Defect triples.}
\label{fig:defects}
\end{figure}

In the following, a modeling approach for microstructural entities, and the evolution of defects and damage therein is proposed. A visualization is provided in Figure \ref{fig:defects}. The concept of grain boundary is modeled in a way where an mm:GrainBoundarySegment is considered to be a portion of the interface delimiting two grains which is approximately planar. The set of all planar segments delimiting two grains is equated to the mm:GrainBoundary class. An mm:GrainBoundaryNetwork spans multiple grain boundaries. We opt for this partitioning approach as it prospectively allows comprehensive modeling of all five macroscopic grain boundary parameters and a description of the grain boundary network's hierarchical structure \cite{saylor2003distribution}. The transitive \textit{mm:isPartOf} object property is used to model this hierarchical structure. 

\begin{figure}[ht]
\centering
\fontsize{7}{10}\selectfont
%\includesvg[width=\linewidth]{images/damage_material_mechanics_ontology_adjusted.svg}
\includegraphics[width=\linewidth]{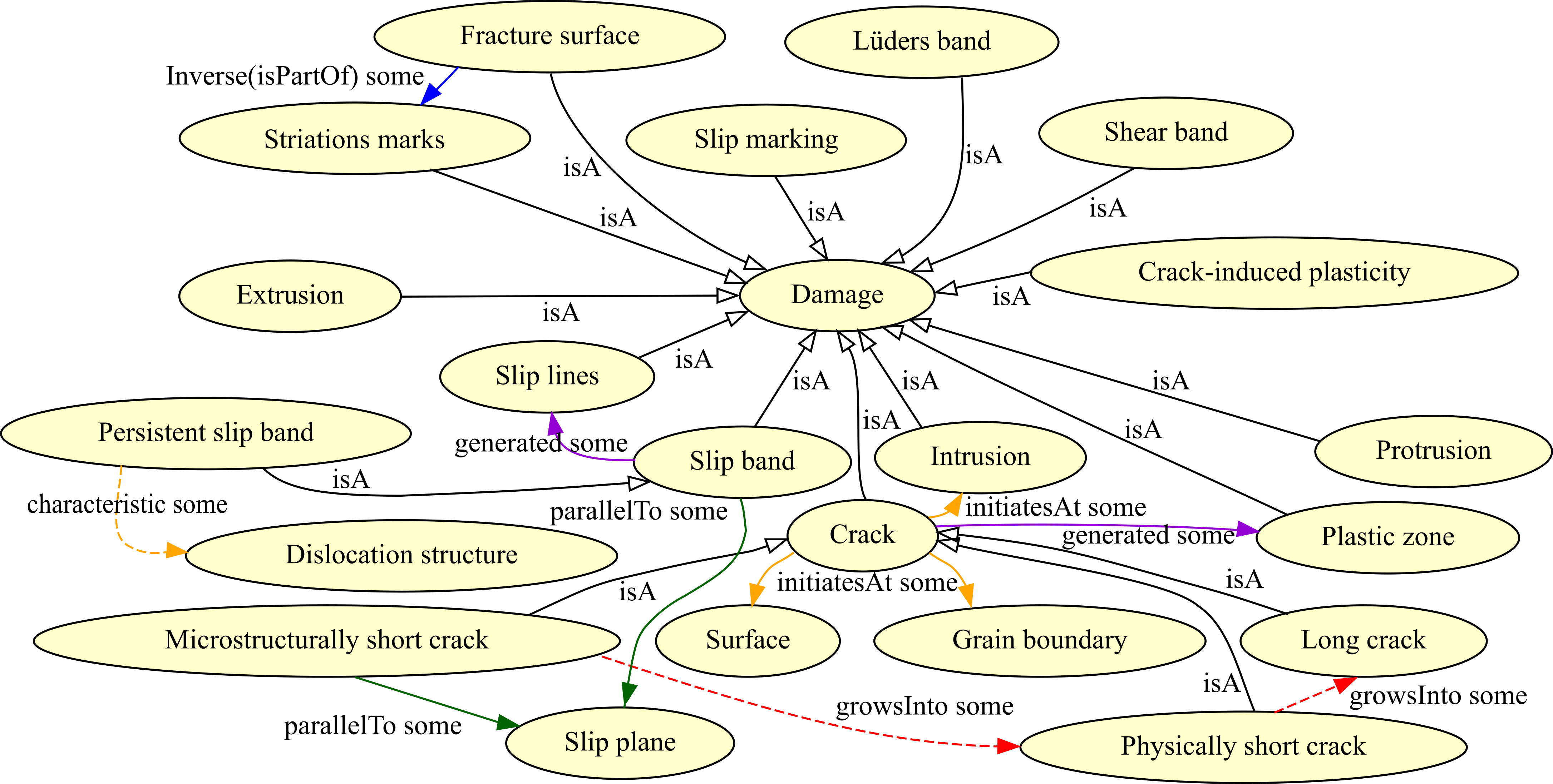}
\caption{The figure shows the ontological modeling of damage including some contextual information. The model covers concepts ranging from local small-scale plasticity to macroscopic cracks and the relations between them. For instance, the fact that slip bands typically are aligned with slip planes is described. Some relations indicate the evolution of cracks from microscopically short over physically short to long cracks, which can be distinguished by the active cracking mode and the plastic zone surrounding the crack. This is modeled after \cite{mcdowell2010microstructure}.}
\label{fig:damage}
\end{figure}

A distinctive factor between the mm:Defect and mm:Damage classes is that crystallographic defects are present at all times of the material life cycle, while mm:Damage is strongly associated with the operation phase. More importantly, individual dislocations are typically considered defects while dislocation structures that formed as a consequence of mechanical loading and are largely irreversible are considered damage \cite{mughrabi2009cyclic}. The concepts that we used for modeling damage are summarized in Figure \ref{fig:damage}. A damage instance can be associated with an initiation site, say a non-metallic inclusion or grain boundary, using the \textit{mm:initiatesAt} object property. The spatial position of entities and activities can be otherwise annotated using the generic \textit{prov:atLocation} object property. Other relevant geometric relations are \textit{mm:alignedWith}, which has a symmetry constraint, and its more specific sub-property \textit{mm:parallelTo}, which can for instance be used to model the characteristic of slip bands to form parallel to twin boundaries in Ni-based superalloys \cite{charpagne2021slip}. As the name implies, \textit{mm:growsInto}, which is a sub-property of \textit{prov:wasDerivedFrom} can be used in cases where an entity grows into an entity of another type. A materials science-specific example would be the growth of a microstructurally-short crack into a physically-short crack \cite{mcdowell2010microstructure}. 

Many of the aforementioned relationships are strong abstractions of the materials domain. The intricate relationships prevalent in many scientific domains call for more expressive extensions of RDF such as the RDF-star standard. The latter standard permits statements about statements. For instance, it could be desirable to add a linear correlation coefficient to the \textit{mm:correlatedWith} object property. That way, complicated reification strategies that are necessary to express the same in RDF can be avoided.

\subsection*{Literature corpus selection}

The literature corpus for the dataset comprises four peer-reviewed open-access publications \cite{durmaz2021efficient, seita2015dual, zhang2020training, geilen2020influence} provided under the Creative Commons CC-BY 4.0 license. The articles center around the microstructure engineering and fatigue domain and also span adjacent domains such as crystallographic defects, characterization techniques, and data processing. Materials fatigue was picked as a domain of interest as mechanistic knowledge for many related phenomena is incomplete and controversially discussed. Moreover, fatigue testing is inherently expensive due to long testing times, making full utilization of information and data from literature all the more important. The dataset covers many of the most relevant alloy categories in engineering, namely steel, aluminum alloys as well as nickel-based superalloys, see the wordcloud in Figure \ref{fig:wordcloud}a for details.

\subsection*{Conversion from Portable Document Format to text}

While some journals offer XML-formatted articles for more effective processing, most of today's literature becomes available in Portable Document Format (PDF). In this work, we have used IBM's Deep Search platform to extract relevant text from our corpus \cite{Staar2018}. The details of the PDF conversion pipelines and models are provided in various publications\cite{Auer_2022, livathinos2021robust, lysak2023optimized, Pfitzmann_2022} and are briefly summarized here. The Deep Search platform can robustly extract texts, images, and tables from both kinds of PDF files (programmatically created or scanned). Deep Search employs a suite of models that are trained to extract bounding boxes or cells for each individual text cell in a PDF. These individual cells are generated by PDF printing commands. Thus, the Deep Search conversion works directly on the native representation of the PDF text cells. Next, the cells are classified into several labels including title, author, abstract, affiliation, subtitle, text, formula, table, image, caption, footnote, citation, keyword, etc. The trained models include YOLOv5 (for object detection) and TableModel (consisting of seq2seq with uni-directional LSTM layers and attention) to accomplish these tasks and achieve state-of-the-art performance (avg. F1 score 0.97) \cite{livathinos2021robust}. Finally, the information contained in the cells and their labels are assembled together in a structured format (json). 

The architecture of the Deep Search platform consists of four layers. There is a frontend layer consisting of REST-API and user interface, which communicates with the next layer. The Orchestration layer consisting of a message broker and a results back-end, schedules all tasks for micro-services, stores their execution status and the final result. Next, there is a compute layer consisting of asynchronous workers executing various micro-services (parsing, predicting, assembly, etc.). This layer is responsible for running models for optical character recognition (OCR), document layout, table structure, reading order, etc. This layer is dynamically scaled and can process about 300 pages parallelly \cite{Auer_2022}. Finally, there is a storage layer that stores all the document and their converted output in an object store and a query-able NoSQL database.

For the dataset at hand, we utilized the textual data contained in abstracts, paragraphs, and figure captions and disregarded any forms of infographics. The text was exported from the Deep Search Toolkit as a structured json document. Paragraphs were subsequently segmented into sentences using the Python package spaCy v3.5.3. This segmentation utilized a pre-trained model `en\_core\_web\_sm' which used the datasets OntoNotes5\cite{weischedel2013ontonotes}, WordNet\cite{miller1995wordnet}, and Constituent-to-Dependency Conversion\cite{choi2015depends} for training, i.e. various genres of text such as blogs, news, and comments. The model achieves a satisfactory performance (F1 score slightly exceeding 0.9 on aforementioned sources) but occasionally fails when special scientific notations occur that include typical sentence delimiters. The new text file which contains line breaks after each new sentence is then imported into INCEpTION\cite{tubiblio106270} for annotation.

\subsection*{Text Annotation}
\label{sec:Annotation_approach}
The following text describes the annotation procedure followed for the creation of the benchmark data. It also can be seen as a guideline if dataset extension is intended. The annotation was performed by three domain experts. Generally, the annotation was performed on the plain text files. In case of ambiguity or uncertainty, the PDF full text and additional infographics therein were consulted. For annotation, relevant named entities in a literature corpus were identified and classified using the INCEpTION annotation tool\cite{tubiblio106270}. INCEpTION has a knowledge base (KB) module that can be used to either create a KB from scratch or upload a file in the resource description framework (RDF) format. The RDF-serialized ontology before PMDco mapping was uploaded in this module. Concepts from the ontology can then be used for the NER annotation. Subsequently, the annotation properties used in the ontology to link concepts to labels were specified in the KB configuration. The "Named Entity" layer in INCEpTION is then configured to have the ``identifier'' field take values from IRIs of classes defined in this KB. From INCEpTION, then files in WebAnno TSV format \cite{TSVWebAnno} were exported. The structure of this file type is explained in section \ref{sec:datarecords}.

\subsubsection*{General annotation approach}

The annotation was done on a fine-granular level, e.g., the term ``edge dislocation'' was annotated as the equivalently named class but then propagated upwards in the taxonomy to the coarse-granular level, i.e., mm:Defect in this case, see section Data post-processing. This was done to obtain a manageable amount of NER classes whilst prospectively permitting subsequent disambiguation of ontological subclasses utilizing the class hierarchy. To annotate an entity, the entire span of a named entity was selected and assigned the appropriate ontology class IRI in INCEptTION. Generally, entities with more than one token can embed sub-entities. We followed the approach by grobid-NER where only the largest entity mention and not the sub-entities are annotated \cite{Grobid}. This results in annotations that are non-overlapping, i.e. flat and not nested. Only when crucial meaning would be lost otherwise or if no appropriate concept was available for the largest entity mention, adjoining entities were assigned separate annotations. In case of ambiguous terms or homonyms, the context was considered to determine the correct annotation, e.g. the term ``fatigue'' can have an emphasis on the experiment or the mechanism. In case uncertainty or ambiguity remained even after reviewing infographics in PDF full-texts, annotations were left unassigned rather than making an incorrect assumption. Common nouns, pronouns, and references that do not represent specific entities were not annotated. Abbreviations or acronyms, were labeled based on their expanded form and separately at the time of introduction. An exception to this are embedded abbreviations such as ``stress-life (S-N) curve'' which were collectively annotated as a single entity.

\subsubsection*{Domain-specific annotation approach}
The Ontology Development section in Methods provides an overview of the domain model's perspectives and consequentially should be considered as a part of the domain-specific annotation guidelines. 
\begin{enumerate}
%\item \textbf{Physical quantities vs. properties:}
\item \textbf{Physical quantities, symbols, and values:} All numeric values, along with units, were annotated as mm:Value. This applies also to hardness values such as ``50 HRC'' in which HRC is also an abbreviation for ``hardness Rockwell''. Symbols such as ``da/dN'' were also annotated as mm:CrackGrowthRate. When ``critical resolved shear stress'' or similar occurred in the text which was not captured in the taxonomy, the whole entity was annotated as mm:Stress, rather than annotating only ``stress''. This also applies to ``maximum''/``mean'' or any other statistical specifiers. Moreover, Miller indices describing crystallographic directions or planes, e.g. \hkl<111>, were annotated as mm:Value. 
\item \textbf{Mechanisms:} The mm:Mechanism class is used for a wide range of entities embodying mechanisms and events such as ``diffusion'', ``aging'', and even ``crack initiation'' or ``high-cycle fatigue''.  
\item \textbf{Materials taxonomy:} All alloy designations, such as ``100Cr6'', ``EN 1.3505'', associated with a defined chemical composition were annotated as mm:Alloy. Whenever an open compound word occurred e.g., ``100Cr6 alloy'', the whole n-gram was assigned the mm:Alloy class. Broader terms such as "aluminum alloy", "stainless steel", "metals" and even "100Cr6 family" etc. were annotated as mm:MaterialClass. For non-metallic materials with defined composition, the mm:Material concept was introduced.
In alloy designations such as ``50CrMo4'', chromium and molybdenum symbols are not annotated as mm:ChemicalElement as this would culminate in nested annotations. Instead, single elements were annotated as mm:ChemicalElement. 
\item \textbf{Materials state descriptor:} In materials science literature, often materials, processes, or specimens and their parameters are jointly described in a compounded and abridged form to distinguish multiple states, e.g. 100Cr6 homogenization heat treatment at a process temperature of 1050 °C as ``100Cr6-HHT1050''. Such constructs were collectively annotated as mm:StateDescriptor. Another common example of this is ``as-received'' or ``underaged''. For instance, for the chunk ``underaged 100Cr6'',  ``underaged'' was annotated with mm:StateDescriptor and ``100Cr6'' with mm:Alloy.
\item \textbf{Positional descriptor:} Often specific locations need to be described, e.g., the crack initiated at the ``surface of the welding seam'', such constructs were annotated as mm:Location.   
%\item In the case of constructs such as '$\alpha$-grain', '$\alpha$' was annotated as "PhaseEntity" and "grain" as "CrystallographicEntity". 
%\item  when there was a shape-specifier before an entity the shape-specifier was annotated jointly, e.g. “unnotched specimen”  
\end{enumerate}

\subsection*{Curation process}
To identify a common ground between the three domain expert raters and improve the data quality, a curation process was implemented. The interrater reliability was computed using the Fleiss kappa value \cite{fleiss1971measuring} on the token level for both label hierarchies, see section Technical Validation. The published dataset is a curated version that aggregates the most appropriate labels from among all three raters. The rater with the most annotation experience acted as a curator. During the curation process, inadvertent mistakes were rectified and annotation consistency was improved.

\subsection*{Data post-processing}
\label{sec:post-processing}
Having the data connected to the ontology permits performing the token classification at various hierarchy levels to alter the number of named-entity types and the semantic gap between them. We have manually selected 27 classes from the ontology as coarse-granular entity types which cover a broad range of entities interesting to materials mechanics data. Each curated annotation, where a text entity is assigned one among 427 classes was propagated to one of its superclasses which is included in the selected 27 entity types. This label propagation process is elaborated below.

% ontology propagation
The annotated data is exported from INCEpTION in WebAnno TSV format, which contains annotations referring to IRIs of the fine-granular ontology entities. The ontology IRIs in the TSV files are used to find the classes they specify from the ontology file. For the fine-granular NER dataset, all annotations are kept as is and are not propagated. In the coarse-granular case, if the annotated class is one among the coarse-granular entity types, the corresponding annotation is kept unchanged. Otherwise, a search is performed upwards in the taxonomy using the \textit{rdfs:subClassOf} object property which terminates when the first class is found that is included in the entity types. The annotation is then assigned this found class. For example, the entity ``edge dislocation'' was annotated as the equivalently named class but since it is not included in the coarse-granular entity types, the annotation is propagated upwards in the taxonomy to the entity type mm:Defect which is included in the entity types.

%BIO tagging + tagset
The mapped annotations are then formatted as a list of entity chunks and their annotated classes in the BIO format (see section Data records). A tag set was created for the coarse-granular dataset containing abbreviations of the 27 low-granularity entity types. For example, mm:Mechanism is abbreviated as MEC, mm:EnvironmentalTestParameter as ETP, and so on. For the fine-granular case, no abbreviations were used but the part of class IRI that contains the name of the class in upper camel case is used as the tag. The BIO tagging scheme is then used where each entity chunk could be assigned either a Beginning, Inside, or Outside tag. A 'B-' prefix followed by the tag from the tagset is attached to the first chunk while the 'I-' prefix is assigned to all chunks that follow and are part of the same entity. The letter 'O' denotes that the chunk is not annotated to an entity type. 

% Final dataset
The most frequent 27 fine-granular named entities in the test set are shown in Figure \ref{fig:Class_distribution}a. Figure \ref{fig:Class_distribution}b shows the 27 distributions in the coarse-granular entity types after the mapping is performed. Because most sentences have only a smaller fraction of chunks that contain the entities we look for, the ’O’ tag predominates in the dataset but is excluded from the figures for visualization purposes.

Once the mapping is complete, the final NER datasets are exported as CoNLL2002 format files.

%\subsubsection*{Third-level section}
 
%Topical subheadings are allowed.

\section*{Data Records}
\label{sec:datarecords}
%The Data Records section should be used to explain each data record associated with this work, including the repository where this information is stored, and to provide an overview of the data files and their formats. Each external data record should be cited numerically in the text of this section, for example \cite{Hao:gidmaps:2014}, and included in the main reference list as described below. A data citation should also be placed in the subsection of the Methods containing the data-collection or analytical procedure(s) used to derive the corresponding record. Providing a direct link to the dataset may also be helpful to readers (\hyperlink{https://doi.org/10.6084/m9                                                                .figshare.853801}

%Tables should be used to support the data records, and should clearly indicate the samples and subjects (study inputs), their provenance, and the experimental manipulations performed on each (please see 'Tables' below). They should also specify the data output resulting from each data-collection or analytical step, should these form part of the archived record.

The dataset and ontology are released under the Creative Commons Attribution 4.0 (CC-BY 4.0) license. The data and codes are maintained on the \href{https://gitlab.cc-asp.fraunhofer.de/iwm-micro-mechanics-public/datasets/materio-miner}{MaterioMiner Gitlab repository}. This publication refers to the release 1.0.0 of the dataset. A static version is additionally hosted with a digital object identifier (10.24406/fordatis/329) on the Fraunhofer research data repository \href{http://dx.doi.org/10.24406/fordatis/329}{Fordatis}. In future dataset releases, we aim to add relation extraction labels --- presumably by using the object properties introduced in the ontology method section. Furthermore, increasing the data quantity is part of our roadmap. This could be achieved by adding more articles to the corpus or by mapping other NER datasets in the MSE domain onto the proposed ontology for joint usage. We invite collaborators to join these efforts. The ontology is available from the \href{https://gitlab.cc-asp.fraunhofer.de/iwm-micro-mechanics-public/ontologies/materials-mechanics-ontology}{Ontology Gitlab repository}. 

\begin{itemize}
\item \textbf{CoNLL 2002 NER files:} In total, 2191 entities of 179 classes from the materials science domain were annotated. The NER data format follows the CoNLL 2002 NER format \cite{tksintro2002conll}, which is an n-column tab-separated text format. In all CoNLL 2002 NER files, the first column is a token and the second column is the NER class. This format applies Beginning–Inside–Outside (BIO) chunk tagging, where the ‘B-’, ‘I-’ prefixes, and ‘O’, are used to indicate the beginning, inside, and outside of an entity, respectively. The end of a sentence is marked by an empty line. The annotations are non-overlapping, i.e. flat and not nested. We provide separate files named `\{article-doi\}.conll' for each annotated scientific article for the \href{https://gitlab.cc-asp.fraunhofer.de/iwm-micro-mechanics-public/datasets/materio-miner/-/tree/main/dataset/fine_granular_ner?ref_type=heads}{FG-NER} and \href{https://gitlab.cc-asp.fraunhofer.de/iwm-micro-mechanics-public/datasets/materio-miner/-/tree/main/dataset/coarse_granular_ner?ref_type=heads}{CG-NER} case. Additionally, we provide three files each for FG-NER and CG-NER where the NER data from the four publications was combined and randomly sampled into training, validation, and testing partitions named equivalently.

\begin{table}[h!]
\centering
\begin{tabular}{|aa|}
\hline
In & O \\
order & O \\
to & O \\
depict & O \\
the & O \\
S-N & B-PDS \\
curves & I-PDS \\
\hline

\end{tabular}
\caption{Structure of the CoNLL 2002 NER format. The left column shows the chunks (in this case words and abbreviations) within a sentence and the right column combines the entity tags and the Beginning–Inside–Outside (BIO) schema. \label{tab:StructureCoNLL}}
\end{table}
\item \textbf{Tag mapping file:} As indicated by Table \ref{tab:StructureCoNLL}, the CG-NER classes are abbreviated and compiled into a tagset, where each tag is composed of three capitalized letters. A mapping of the tags to the ontological preferred labels (rdfs:label annotation property) is given in the comma-separated-value file named 'CG-NER\_tags.csv', where the first and second column represent the tag and the preferred label, respectively.      
\item \textbf{Webanno TSV file:} Furthermore, a WebAnno TSV 3.3 format file is provided which is a tab-separated value (TSV) file commonly used for named-entity linking. It consists of a header and body, where the latter is grouped into sentences. There, each unprocessed sentence is provided along with the TSV-formatted section which features the tokenized sentence, sentence numbers, character offsets, and if applicable, international resource identifiers (IRIs) that link directly to the annotated ontological classes of the ttl-file. For a more detailed description of the format, we refer to \cite{TSVWebAnno}. We provide an `\{article-doi\}.tsv' for each annotated scientific article for the FG-NER setting.
\item \textbf{Ontology file:} The corresponding class definitions and object properties are provided in the accompanying ontology serialized as a turtle file (materials\_mechanics\_ontology.ttl and materials\_mechanics\_ontology\_merged.ttl). The latter file combines the materials mechanics ontology along with its imports into a single turtle file, which can also be imported to an annotation tool, e.g., INCEpTION. We host the ontology on a separate git repository to facilitate collaborating, extending, and versioning.
\item \textbf{Training scripts and utility functions:} We provide Jupyter notebooks that contain the procedure for fine-tuning and evaluation as shown below in the Technical Validation section. Example notebooks named ``ner\_finetuning.ipynb'' to process both fine- and coarse-granular CoNLL files are provided along with utility functions in ``utils.py''. There, the import and processing of the data can be observed. Additionally, a notebook named ``propagate\_labels.ipynb'' that propagates the annotation within the TSV formatted file to a desired tagset is also provided.
\end{itemize}

\section*{Technical Validation}
\label{sec:techval}

%\subsection{Scope and domain of the ontology}
%This section presents any experiments or analyses that are needed to support the technical quality of the dataset. This section may be supported by figures and tables, as needed. This is a required section; authors must present information justifying the reliability of their data.

\subsection*{Annotation consistency}
To assess annotation consistency between the three raters, Fleiss kappa scores were computed. These hold compounded information about the specificity of the ontological description, annotator background, and clarity of the text documents. This metric reaches up to unity for complete agreement and negative scores indicate an entirely random process. These scores were computed for different settings as outlined in Table \ref{tab:Interrater}. Overall the highest Fleiss kappa scores reach a value of 0.733 which is typical in materials science-related NER annotation \cite{yang2022pcmsp, foppiano2021supermat}. However, it is crucial to note that even in the lower granularity case 27 different entity types were considered, while in the high granularity case, there are 427 possible ontology classes to pick from. This not only enriches the information density of the fine-granular annotations but also inevitably narrows down the semantic distance between ontological concepts and opens up different possibilities for annotating the same textual entity. Consequently, this can render the annotation more challenging, and a high degree of familiarity with the ontology is required to perform the annotation precisely. As opposed to other works, where multiple annotation cycles over the whole dataset were performed \cite{yang2022pcmsp, foppiano2021supermat}, the dataset presented here was annotated in a single iteration by each rater followed by a curation process through a single person. 

\begin{table}[ht]
\centering
\begin{tabular}{|l|c|c|c|c|}
\hline
& \multicolumn{2}{c|}{\textbf{Coarse-granular NER}} & \multicolumn{2}{c|}{\textbf{Fine-granular NER}} \\
 \hline
& with BIO annotation & w/o BIO annotation & with BIO annotation & w/o BIO annotation \\
\hline
\textbf{Annotated tokens} & 0.572 & 0.608 & 0.504 & 0.538 \\
\hline
\textbf{All tokens} & 0.701 & 0.733 & 0.650 & 0.677 \\
\hline
\end{tabular}
\caption{\label{tab:Interrater} Fleiss kappa scores for inter-rater variability assessment between the three raters for different conditions. The `all tokens' and `annotated tokens' rows take into consideration all tokens and only such that were annotated as an entity of interest by at least one rater, respectively. The coarse-granular NER column refers to the annotations after propagation while fine-granular NER considers the detail labels. Further, we distinguish between two cases, considering the BIO tag or ignoring it.}
\end{table}

\begin{figure}[ht!]
\centering
\includegraphics[width=0.92\linewidth]{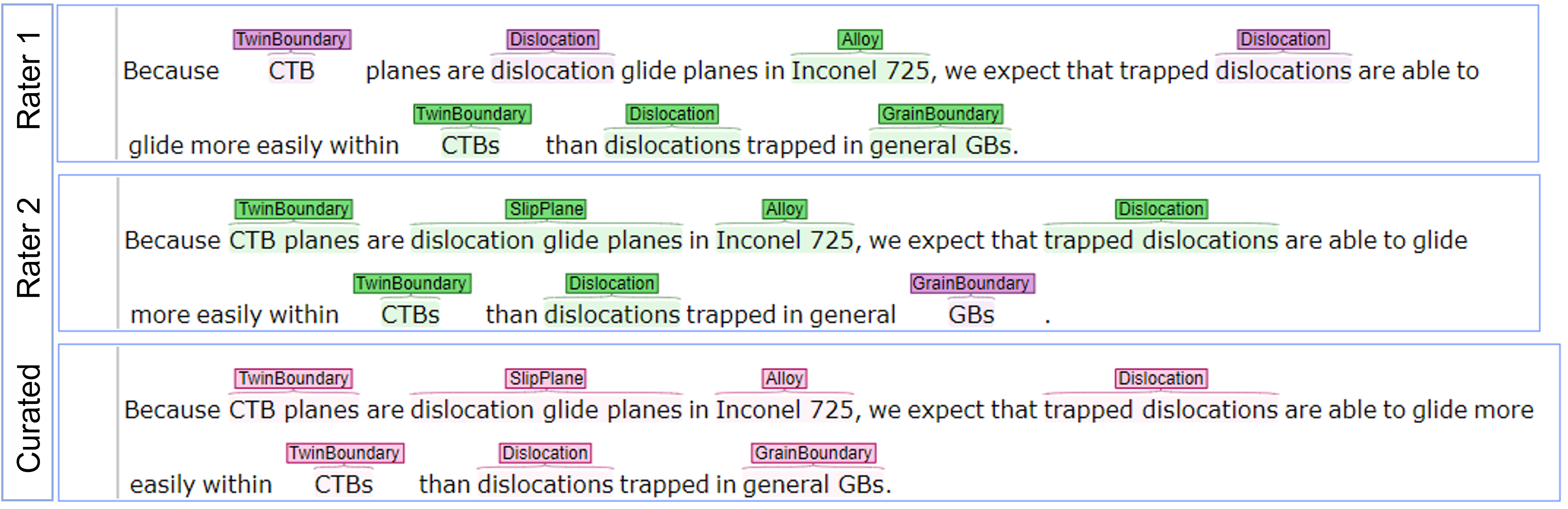}
\caption{Annotation and curation of a sample sentence. All shown annotations are at the entity level. The green, violet, and pink colors in the annotation boxes indicate accordance, deviations, and curated annotations, respectively.}
\label{fig:rater_comparison}
\end{figure}

Some trends can be observed in the Fleiss kappa $\kappa$ scores. The BIO annotation reduced the scores by 0.027--0.036 depending on which tokens are accounted for. This indicates that the raters chose distinct start and end tokens for annotating relevant entities occasionally. A typical example of this is shown in Figure \ref{fig:rater_comparison}, where the first two entities ``CTB planes'' and ``dislocation glide planes'' were annotated distinctly by two raters. While neither of the annotations is incorrect, in this case, the largest entity (see annotation guidelines in section Annotation approach) was selected during the curation stage. Concerning the last entity ``general GB'', we interpreted it as ``non-special GB'', which is a frequently occurring term and important specifier in MSE\cite{shvindlerman1985regions} which is why we opted for including ``general'' in the annotation. When disregarding entities that have been consistently annotated as O-Outside, the impact is substantial ($\Delta\kappa \approx 0.13$). This is expected since the common outside tokens are typically easy to identify as such. Another interesting comparison can be made between the CG-NER and the more detailed FG-NER annotations. On this end, a comparatively small score difference of 0.051--0.07 can be observed. When propagating upwards in the class hierarchy of the ontology, smaller annotation errors of nearby concepts such as twin and grain boundaries are mitigated as both have the same superclass mm:Defect. The influence of such class confusion on inter-rater consistency is stronger than BIO tagging errors (span errors). Allowing nested annotations would alleviate both errors since the raters would not have to decide between adhering to annotation guidelines (e.g. largest entity mention) or preserving most information. Furthermore, adding more descriptive definitions in the ontology can improve the annotation consistency for future extension of the dataset. 

\subsection*{Class distribution}

\begin{figure}[ht!]
    \subfloat[\textbf{Material}]{\includegraphics[width=0.48\textwidth]{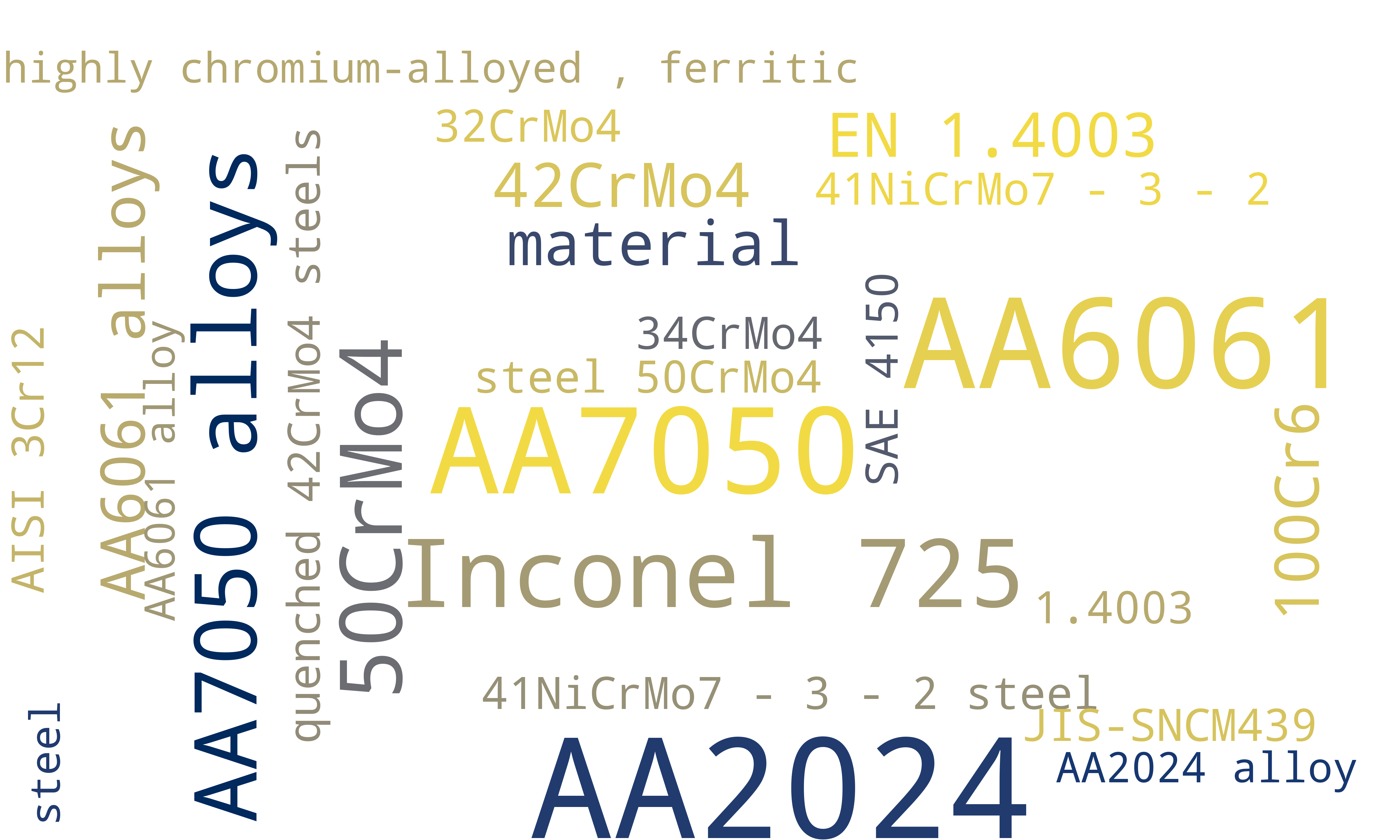}} 
    \subfloat[\textbf{Defect}]{\includegraphics[width=0.48\textwidth]{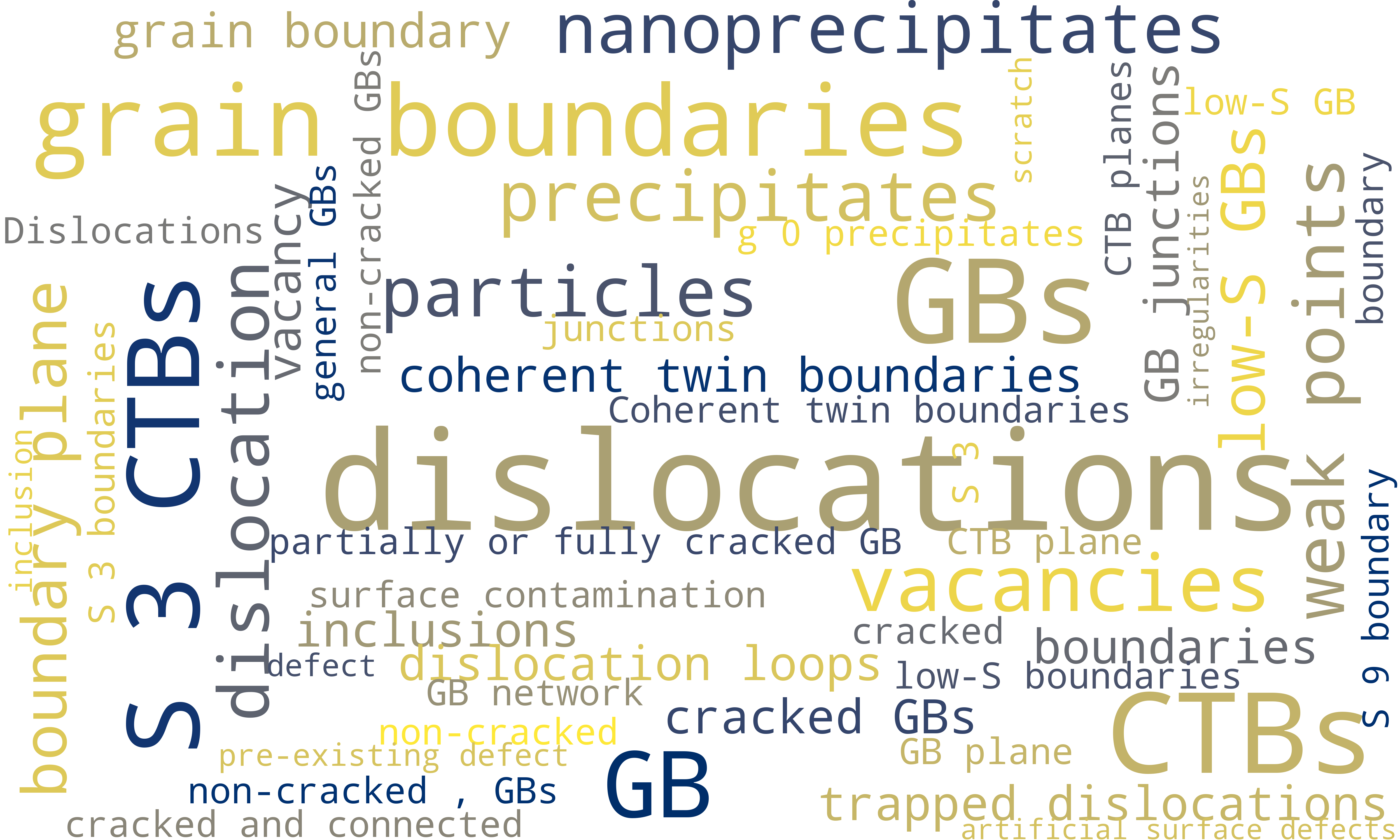}} \\
    \subfloat[\textbf{Mechanism}]{\includegraphics[width=0.48\textwidth]{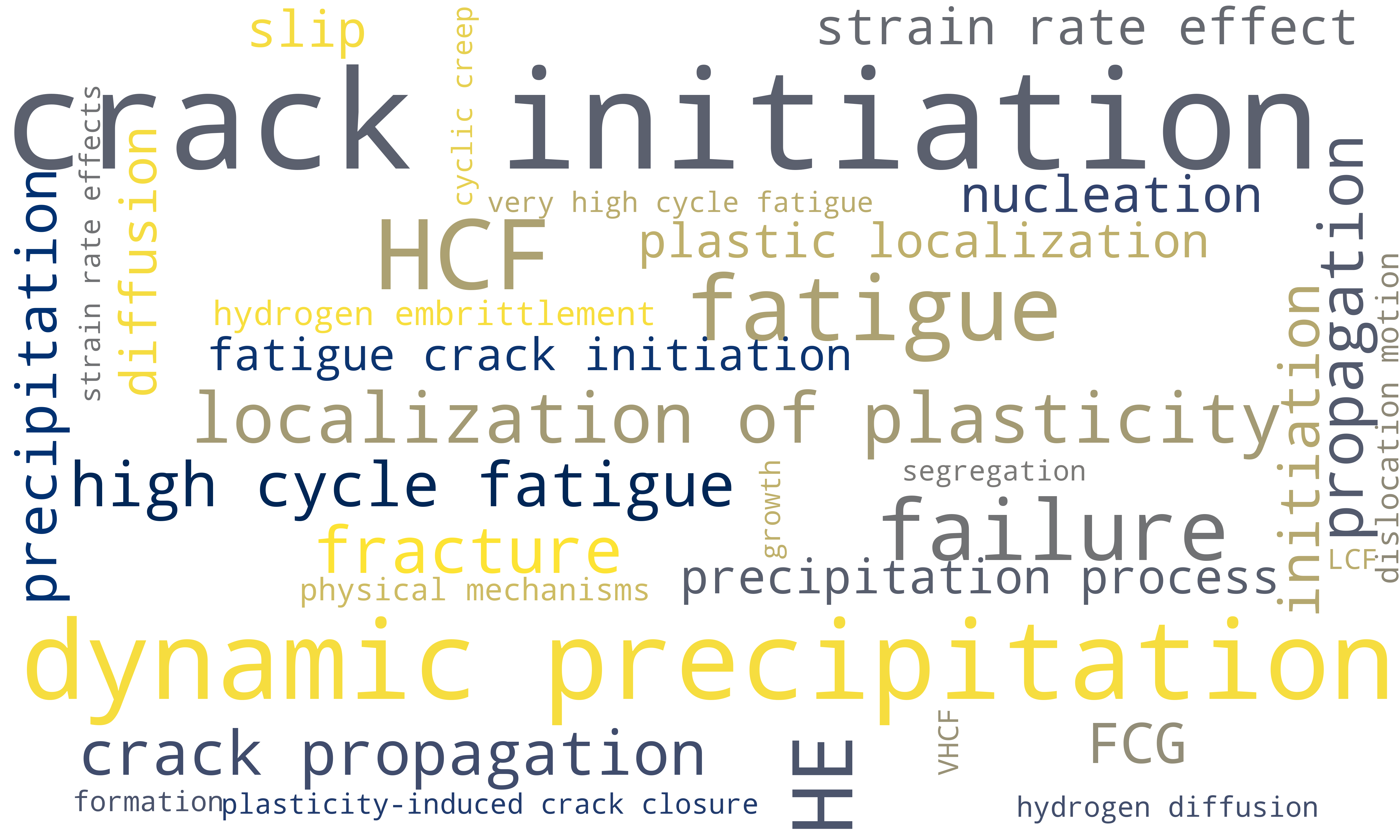}}
    \hspace{4mm}
    \subfloat[\textbf{Characterization Process}]{\includegraphics[width=0.48\textwidth]{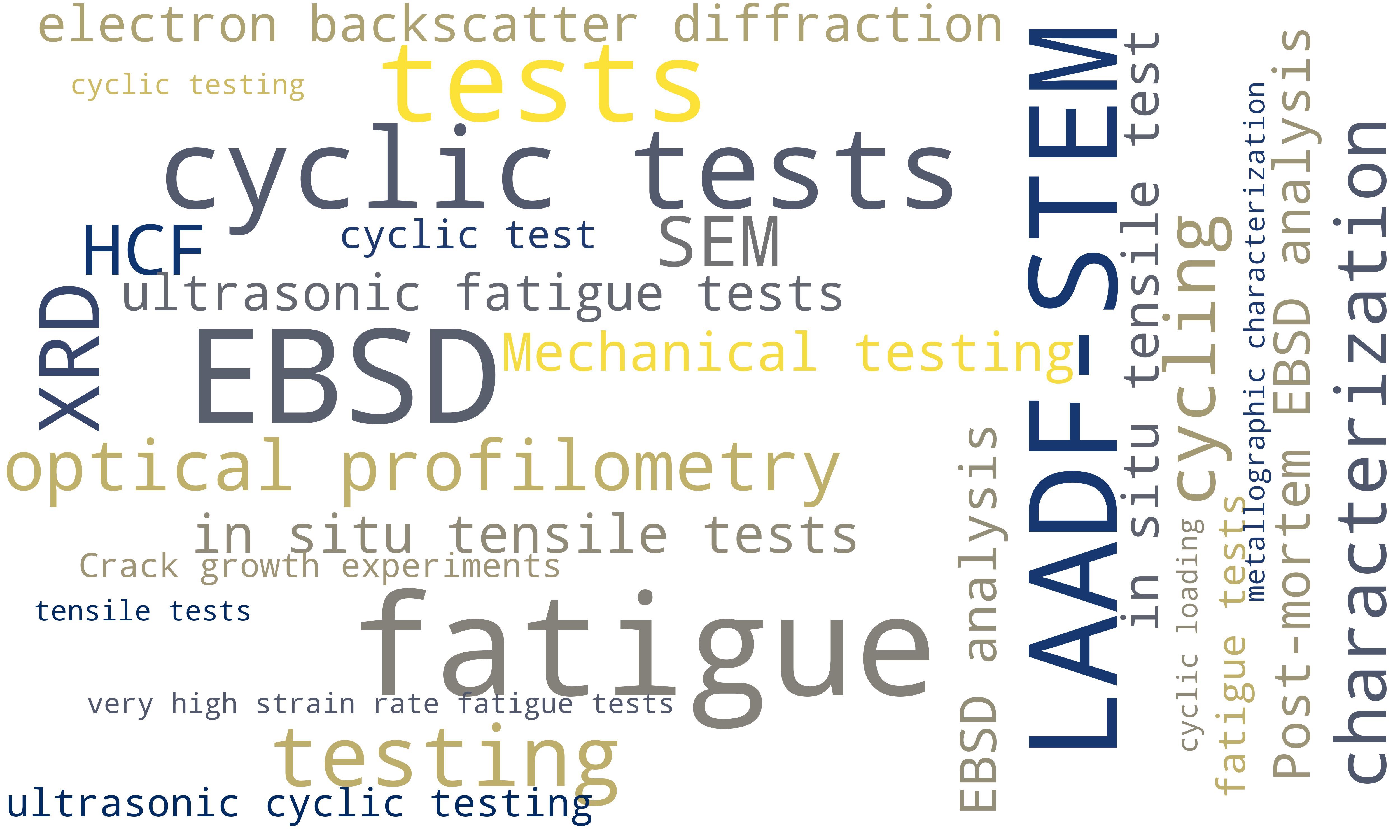}}

\caption{Word clouds indicate the words annotated in the text for the named-entity types mentioned in the subcaptions. The word clouds were generated with the Wordcloud python package \cite{mueller_2023_10321882}. The font size is directly proportional to the frequency of occurrence (setting the relative scaling parameter of the WordCloud function to a value of 0.5).}
\label{fig:wordcloud}
\end{figure}

\begin{figure}
\centering
\subfloat{
   \includegraphics[width=1\linewidth]{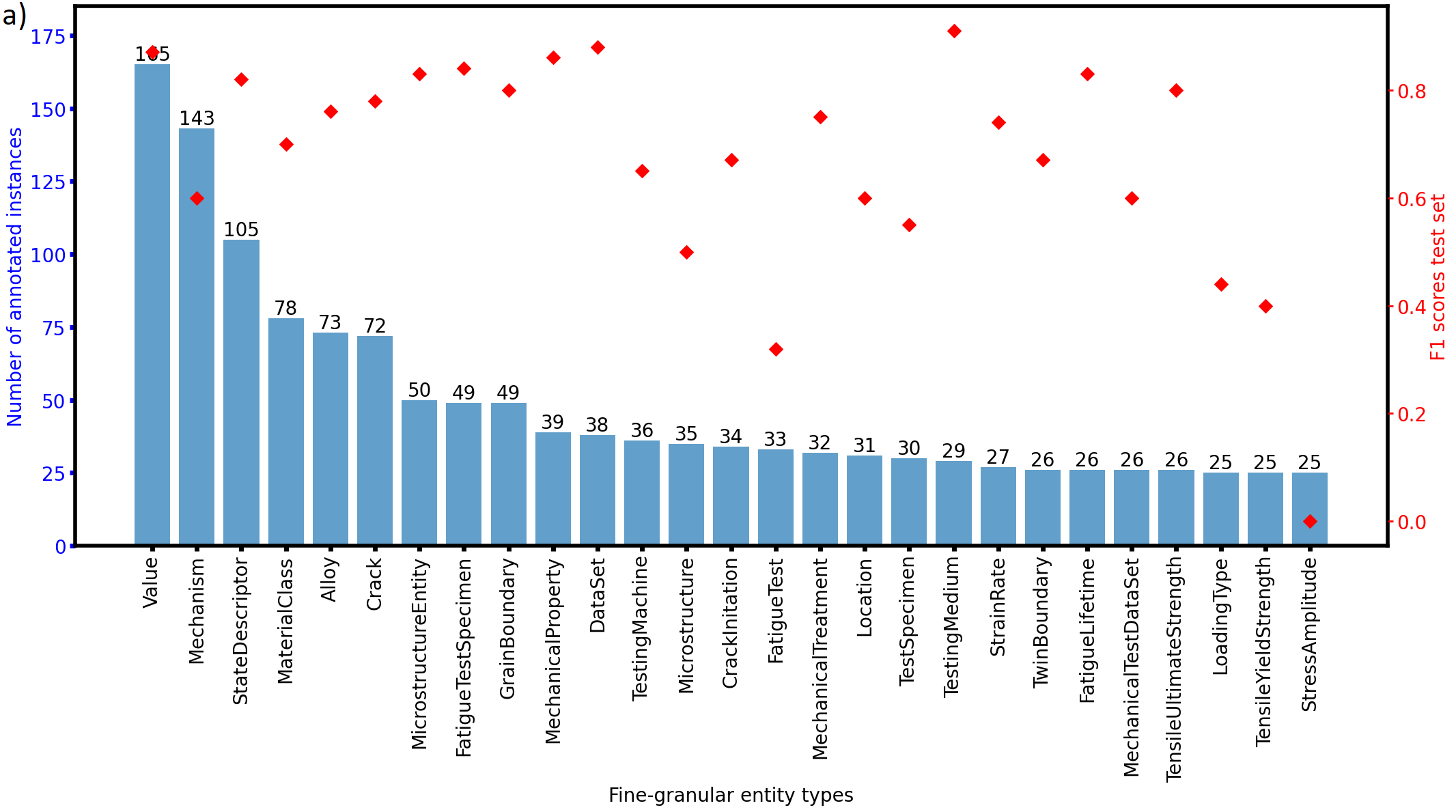}}\\
\subfloat{
   \includegraphics[width=1\linewidth]{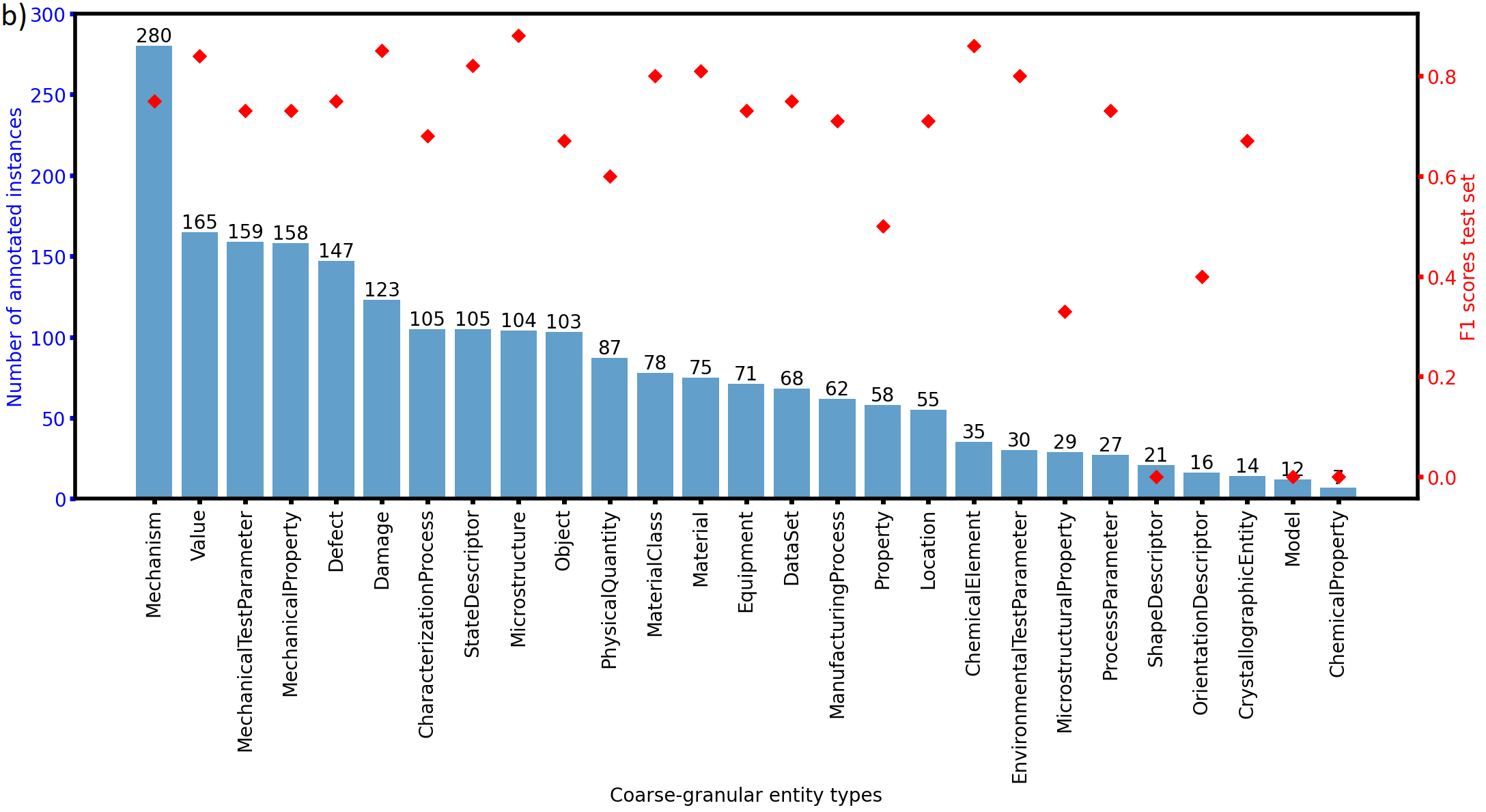}}
\caption{Histograms showing the overall distribution of classes and F1 scores obtained on the test subset. Figure (a) shows the top 27 fine-granular named-entity types and (b) all coarse-granular named-entity types. \label{fig:Class_distribution}}
\end{figure}

In Figure \ref{fig:Class_distribution}, the class distribution within the dataset at the fine-granular (Figure \ref{fig:Class_distribution}a) and coarse-granular level (Figure \ref{fig:Class_distribution}b) is depicted. The definition of the pre-defined high-level classes directly influences the class imbalance and thus the training of the models which in the worst case can lead to biased models. The most frequently annotated FG-NER classes are all largely plausible given the corpus at hand and the distribution seems representative of materials mechanics literature. Some fairly specific classes such as mm:TwinBoundary have higher occurrence than anticipated in MSE, which is owed to the inclusion of the publication of Seita et al. \cite{seita2015dual} describing their influence on crack initiation in great detail. Overall, the fine-granular histogram is comparatively more uniform. This indicates that the clustering of ontological classes, i.e. upward propagation in the taxonomy, adds some additional skew. Entities belonging to the mm:CrackInitiation classes are propagated to the mm:Mechanism class. Thus, in the predefined set of coarse-granular entity types, the mm:Mechanism class is most frequent which is a direct consequence of the class' broad interpretation. While most illustrated disparities in class occurrence are owed to inherent characteristics of the publications, further partitioning of the mm:Mechanism class could alleviate the disproportionately high emergence of this entity type. Furthermore, unsurprisingly, values were encountered often, closely followed by mechanical test parameters and properties. This is a result of corpus selection where mechanics of materials publications were hand-picked. Another selection criterion was that different defects and their influence on materials properties and degradation behavior are represented which is shown in the mm:Defect and mm:Damage class occurrences. In Figure \ref{fig:wordcloud}, we show the annotated entities for the four classes mm:Materials, mm:Defect, mm:Mechanism, and mm:CharacteriztionProcess within word clouds where the font size correlates with the frequency of occurrence. One can observe that a variety of relevant entities in materials mechanics are covered despite limiting the dataset to four publications.

\subsection*{Validation through NER model training}
To assess whether the dataset permits the supervised fine-tuning of specialized NER models, we perform a standard fine-tuning on top of the pre-trained MatSciBERT model \cite{gupta2022matscibert}. This allows us to assess whether available data quantity is suitable for model fine-tuning or whether it can only be used in the context of few-shot learning and prompting as well as benchmarking. 

Therefore, a random data split with 65\% training, 15\% validation, and 20\% test data was performed on both distributions showcased in Figure \ref{fig:Class_distribution}, i.e., both the coarse and fine-granular NER tasks. The sentences were sampled randomly such that we obtained roughly the same class distribution for the three subsets. In the case of FG-NER, we opted for using the top 27 most frequent classes (shown in Figure \ref{fig:Class_distribution}) instead of all available 179 annotated classes. Then the training took place using the Hugging Face transformers package \cite{wolf2020transformers}. The model initialization to the pre-trained MatSciBERT model was done using the model option `m3rg-iitd/matscibert'.

Data augmentation was disregarded. A simple cross-entropy loss and AdamW optimizer were applied. Hyperparameter tuning was performed for the learning rate, weight decay, and batch size. Ultimately, for training, the AdamW optimizer with settings $\beta_1 = 0.9$, $\beta_2 = 0.999$ and $\epsilon = 1.1^{-8}$ and a batch size of 32 was employed. Each instance was based on isolated sentences, i.e., without providing additional context. The evaluation of the model was performed using the seqeval python package\cite{seqeval} and the `classification\_report' function contained therein, which also provides class-resolved F1 scores. Both the overall and class-resolved F1 score computations were performed at the entity level considering both type and span matches.

After fine-tuning, overall test F1 scores of 69.92\% and 72.32\% were obtained for the FG-NER and CG-NER tasks, respectively. The validation scores were slightly higher, namely 72.25\% and 74.48\% each. These performances are satisfactory and indicate that a model can be fine-tuned using a subset of the published dataset. Both F1 scores are similar which indicates that the propagation does not introduce major inconsistencies and that the intra-class variance is still manageable for the CG-NER model. The marginal increase of CG-NER can probably be attributed to the surplus of data (in the FG-NER case infrequent classes were dropped rather than propagated upwards). Moreover, a stronger semantic separation between the classes might contribute to this observation. Since the whole data set was drawn from four publications, the out-of-distribution generalization of the model can not be estimated. The performance would probably fall short if different MSE subdomains, wording, and writing styles were concerned. The class-wise performance is illustrated on the second y-axis of Figure \ref{fig:Class_distribution}. One can observe that the F1 score falls slightly short for the minority classes. This can presumably be alleviated by introducing a weighting term based on the inverse class frequency in the loss function. 

\section*{Usage Notes}

%The Usage Notes should contain brief instructions to assist other researchers with reuse of the data. This may include a discussion of software packages that are suitable for analyzing the assay data files, suggested downstream processing steps (e.g. normalization, etc.), or tips for integrating or comparing the data records with other datasets. Authors are encouraged to provide code, programs or data-processing workflows if they may help others understand or use the data. Please see our code availability policy for advice on supplying custom code alongside Data Descriptor manuscripts.

%For studies involving privacy or safety controls on public access to the data, this section should describe in detail these controls, including how authors can apply to access the data, what criteria will be used to determine who may access the data, and any limitations on data use. 

\begin{itemize}
    \item \textbf{Literature corpus extension:} When further scientific articles or other textual sources are supposed to be supplemented, either direct PDF annotation in INCEpTION can be performed or the PDF files can be converted to machine-readable formats through Grobid, Deep Search, or another conversion tool. The open-sourced Deep Search Toolkit provides an easy interface to convert single or multiple PDFs. In addition, the Toolkit offers additional capabilities like extracting chemical entities (text or figure), a vast library of documents, and document question-answering\cite{mishra2023esg}. When Grobid is used, the resulting TEI-XML files can be processed with packages that handle XML files, such as BeautifulSoup \cite{richardson2007beautiful}. Different relevant metadata and paragraphs can be extracted. Note that the underlying data-driven models of these conversion pipelines often fall short when documents with non-standard format are concerned. 
    \item \textbf{Corpus extension by integration with other existing NER datasets:} One desirable objective could be to harmonize and merge the NER datasets listed in Table \ref{tab:dataset_comparison} to increase the data quantity and scope. In that case, annotated classes in the other datasets (see the last row of Table \ref{tab:dataset_comparison}) could either be equated to existing classes within the published ontology or added as new concepts. Especially when an equivalence mapping is to be performed the accordance between our ontological class definition and their annotation approach should be carefully reviewed. Based on this ontology mapping approach, the BIO tags could then be updated and assigned. The varying degree of sparsity in annotation, as shown in Table \ref{tab:dataset_comparison}, should be considered for model training.
    \item \textbf{Data annotation and correction:} The CoNLL 2002 NER and TSV web anno files can be imported to INCEpTION to adjust or extend the annotations. This requires that the named-entity annotation layer in INCEpTION is configured to use the materials mechanics ontology as the knowledge base and the tagset corresponding to the CoNLL file as the tagset. Relation extraction annotations can be added that way as well given a relation layer is configured. The TSV files could also be used to create CoNLL files with a different tagset by following the label propagation steps described in section \ref{sec:post-processing}. The code for performing this is provided in the Gitlab repository. 
    \item \textbf{Ontology display and manual adjustments:} For a glance into the ontology, we recommend using the \href{http://matportal.org/ontologies/MECH}{Matportal ontology viewer} where the ontology is hosted or the \href{https://w3id.org/pmd/materials-mechanics-ontology/1.0.0}{ontology specification page} where all concepts are specified and visualized as a graph using WebVOWL 1.1.7. The \href{https://gitlab.cc-asp.fraunhofer.de/iwm-micro-mechanics-public/ontologies/materials-mechanics-ontology}{ontology git repository} can be forked for further developments. Modifications and extensions to the ontologies can be made using the Protégé application which is the de facto standard in ontology development. There, classes, individuals, object/data properties, or restrictions can be added as required. Protégé also supports reasoning with reasoners like Hermit\cite{glimm2014hermit}, Fact++\cite{tsarkov2006fact++}, and ELK\cite{kazakov2012elk}. Reasoning on the ontology is tested with the Fact++ reasoner. The authors recommend collaborators to follow standard git workflows and use the issues and pull requests in the git repository.
    \item \textbf{Automated ontology processing:} A programmatic way to manipulate the OWL2 ontology is the python package owlready2 \cite{lamy2017owlready}. It provides a means to create, extend, and partition ontologies, and perform reasoning on them. It is integrated with a SQLite3-based triple store which supports some SPARQL query functionality. Thus, owlready2 can prospectively be used to compile knowledge graphs from the textual data. Alternative ways are Rdflib in Python or java-based OWLAPI \cite{horridge2011owl} and the ROBOT package \cite{jackson2019robot} which also permit managing OWL2 ontologies. For instance, these tools permit merging concepts that are agnostic to the specific MSE subdomains into other upper-level ontologies.
    \item \textbf{Training of ML and neurosymbolic models:} When training or programmatic post-processing of the data is concerned, the NER data can be imported into Python by using Spacy, NLTK, HuggingFace, and Pandas open-source NLP packages. After importing typically a subword tokenization step is performed for training or evaluation on the data. To achieve this, HuggingFace provides a library of pre-trained tokenizers. Rather than training NER models such as BERT, Deberta, Roberta, and others from scratch, pre-trained weights can be obtained. Especially for BERT, there are already pre-trained materials science models available providing a good starting point. An example of this procedure is given in the \href{https://gitlab.cc-asp.fraunhofer.de/iwm-micro-mechanics-public/datasets/materio-miner}{MaterioMiner Gitlab repository} as described in the Data Records section. The provided materials mechanics ontology and its underlying graph structure, lexical information, and logical constructors can be used for machine learning by using embeddings generated through OWL2vec* \cite{chen2021owl2vec}.
\end{itemize}

\section*{Code availability}

We provide the source codes for training a language model in the accompanying \href{https://gitlab.cc-asp.fraunhofer.de/iwm-micro-mechanics-public/datasets/materio-miner}{Gitlab repository}. With the provided source code the fine-tuning of the pre-trained MatSciBERT and evaluation can be reproduced.  

%For all studies using custom code in the generation or processing of datasets, a statement must be included under the heading "Code availability", indicating whether and how the code can be accessed, including any restrictions to access. This section should also include information on the versions of any software used, if relevant, and any specific variables or parameters used to generate, test, or process the current dataset. 

%\bibliographystyle{unsrtnat}
\bibliographystyle{plain}
\bibliography{sample}

\section*{Acknowledgements}

The authors affiliated with Fraunhofer Institute for Mechanics of Materials express their gratitude to the German Federal Ministry of Education and Research (BMBF) for funding in the scope of the iBain project (13XP5118B) as part of MaterialDigital.
%Acknowledgements should be brief, and should not include thanks to anonymous referees and editors, or effusive comments. Grant or contribution numbers may be acknowledged.

\section*{Author contributions statement}

A.R.D. and A.T. conceptualized the dataset and its analysis, A.R.D. and A.T. developed the ontology. A.R.D., A.T., and T.S. performed the data annotation and curation, A.T., L.M., and R.N.M. performed the data post-processing and analysis, A.R.D and A.T. analysed the results. All authors contributed to the writing and review of the manuscript. 

\section*{Competing interests}

The authors declare no competing interests.





% \end{thebibliography}

\end{document}